# Automatic Detection of Font Size Straight from Run Length Compressed Text Documents

Mohammed Javed[#], P. Nagabhushan[#], B.B. Chaudhuri[*]

[#]Department of Studies in Computer Science
University of Mysore, Mysore-570006, India
[*]Computer Vision and Pattern Recognition Unit
Indian Statistical Institute, Kolkata-700108, India

*Abstract*— **Automatic detection of font size finds many applications in the area of intelligent OCRing and document image analysis, which has been traditionally practised over uncompressed documents, although in real life the documents exist in compressed form for efficient storage and transmission. It would be novel and intelligent if the task of font size detection could be carried out directly from the compressed data of these documents without decompressing, which would result in saving of considerable amount of processing time and space. Therefore, in this paper we present a novel idea of learning and detecting font size directly from run-length compressed text documents at line level using simple line height features, which paves the way for intelligent OCRing and document analysis directly from compressed documents. In the proposed model, the given mixed-case text documents of different font size are segmented into compressed text lines and the features extracted such as line height and ascender height are used to capture the pattern of font size in the form of a regression line, using which the automatic detection of font size is done during the recognition stage. The method is experimented with a dataset of 50 compressed documents consisting of 780 text lines of single font size and 375 text lines of mixed font size resulting in an overall accuracy of 99.67%.**

## I. INTRODUCTION

Automatic detection of font size in text documents is an important and crucial pre-knowledge extraction stage in applications of intelligent OCRing and document image understanding [1, 2]. Further, in case of document structure analysis, the pre-knowledge of font size is an added advantage in understanding the title, section, subsection, paragraph, etc from a document image[1]. Different applications like phrase spotting[2], bold words detection[1], forgery detection[3], line segmentation[2] have been proposed in the literature that emphasize the detection of font size as key stage in improving the performance of the system. However, these techniques are designed for documents in uncompressed form. On the contrary, in reality most of the document processing systems like fax machines[4], xerox machines, and digital libraries use compressed form to provide better transmission and storage efficiency. Although in real life the documents are made available in compressed form, the existing system has to decompress the compressed document and then operate over it, which requires additional computing resources. Therefore, working on compressed data directly without decompressing has proven to offer lot many advantages in terms of time and space[5, 6, 7] and hence has become a priority area of research.

As our research goal is to make way towards intelligent OCRing and document understanding by automatically detecting font size from compressed text documents, we limit our research to only binary documents. The popular compression technique available for both archival and transmission of text documents is CCITT Group-3[8], which is widely used in the form of TIFF and PDF documents. CCITT Group-3 is supported by wide range of fax machines and hence direct operations on these compressed documents will help in developing intelligent and efficient applications such as word spotting[5], duplicate document detection[9] and retrieval[10]. The backbone of CCITT Group-3 compression algorithm is run length encoding technique. In this backdrop, this research study is specifically focused on direct processing of document images in run length compressed domain.

In the literature, some of the latest works related to run length compressed document processing are feature extraction[7], entropy computation[11], text document segmentation[12] and block extraction[13]. Moreover, to our best knowledge, we could not find any research attempt in detecting font size in compressed documents which is considered to be a foundation work for intelligent OCRing and document analysis using compressed documents. Therefore in this research work, we aim at developing an automated system for learning and detecting of font size of text lines straight from run length compressed TIFF documents without undergoing through the stage of decompression. Rest of the paper is organized as follows: section-2 gives background details of the compressed document from text line perspective, section-3 discusses the proposed model for font size detection, section-4 shows the experimental analysis with the proposed methods, section-5 concludes the paper with a brief summary.

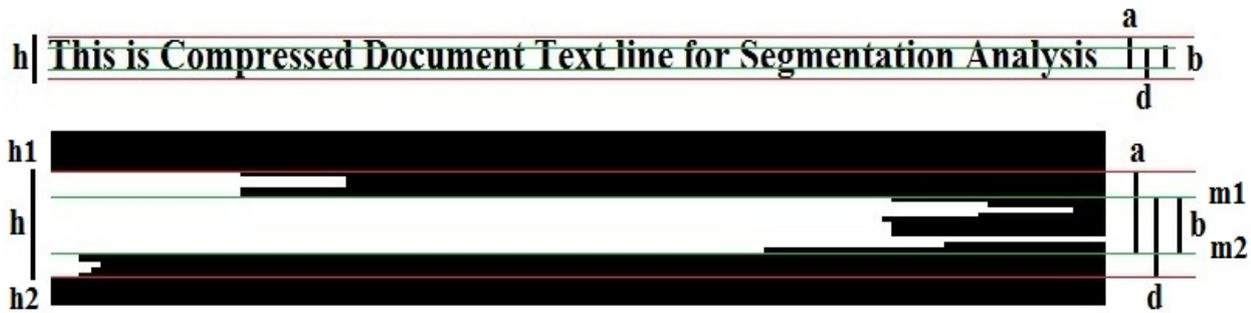

Fig. 1: Height features of a text line in uncompressed and compressed versions ('h'-Line height,'b'-Base height,'a'-Ascender height,'d'-Descender height)

$$r = \sum_{j=1}^{\frac{n'}{2}} \{w(1,j) + b(1,j)\} - \{\{\min\{w(i,1)\} + \max\{w(i,\frac{n'}{2})\}, \forall i = 1..m'\} \qquad (1)$$

## II. LEARNING ABOUT A COMPRESSED DOCUMENT

In this section, we discuss about the different characteristic features associated with a compressed text document specifically from its text line perspective. We brief here various issues governing a compressed text line, their related density parameters followed by a brief introduction of regression based learning approach for font size detection.

### A. Text Line Parameters

Any text document image can be described at different resolution levels such as page, paragraph, line, word and character. At all these levels it is possible to have text consisting of various font size and font style. Also, we know that most of the text in a document is usually characterized in terms of font size, therefore there is a need to develop a system that keeps track of the structural details of the text during OCRing which may be called as intelligent OCRing of texts for applications in editing and reprinting of documents as emphasized by [1, 3]. This paper is an attempt to support intelligent OCRing which aims at developing an automated system for preserving the structural details of compressed text lines through automatic detection of font size directly from compressed documents.

Most of the time, text lines in digitally archived documents such as articles, books, newspapers, magazines come with single font style and multiple fonts size. Therefore detecting the font size of the text lines for a given font style is an important and intelligent task before OCRing of these documents. In any font style, generally a text line consists of three categories of text: uppercase, lowercase and mixed-case. Among them, the mixed-case types of text lines are more common in practical life. Therefore, in this research we aim to demonstrate font size detection with mixed-case text lines of Arial font. A text line sample for a mixed-case text is shown in Fig-1 in both compressed and uncompressed versions, where the text line parameters such as line height(h), base height(b), ascender height(a), descender height(d) are defined. These mixed-case text lines can be further classified into three types:(i) text line with both ascenders and descenders, (ii) with ascenders or descenders and (iii) without ascenders and descenders. However, one should be aware that, it is less probable to get a long text line without any ascender characters in them[2].

The run length compressed data of the text line in case of binary documents consists of alternate columns of white runs (w(i,j)) and black runs (b(i,j)). Let this compressed data of the text line be denoted in two column matrix format as L(w(i,j),b(i,j)) where i = 1..m' and j = 1..$\frac{n'}{2}$ . Here m' and n' respectively denote number of rows and columns present in the compressed text line. The representation L(w(i,j),b(i.j)) for the compressed text line provides access to two columns at a time, therefore the number of columns in the compressed text line becomes $\frac{n'}{2}$ .

Another important parameter of the compressed text line is its length, which is equal to number of columns in the compressed data of the text line. Let this length be denoted as 'l' and its value given as l = n'. From the compressed text line it is also possible to compute the length of the original text line in the uncompressed version and let this length be denoted as 'r' which can be computed with equation (1).

In text lines of different font size, there is an interesting pattern observed in the length of compressed text lines. When the font size of the text line increases, the length of the line in the compressed version decreases gradually which can be seen clearly in Fig-2. Such a pattern is observed because of increase in stroke width of the characters with increase in font size, which naturally reduces the total number of runs in a text line of fixed length and the line length gets shortened with increase in font size. The length pattern for different font size shown in Fig-2 is observed for text lines of full length. However in case text lines having their length less than full length, the length pattern of the text line may overlap with the existing length pattern of other font size. Therefore in order to use this length as distinguishing feature for any application, we recommend to normalize this length(l) with the original length(r) of the text line which may be called as normalized length ratio(R) and is mathematically expressed as

$$R = \frac{l}{r} \qquad (2)$$

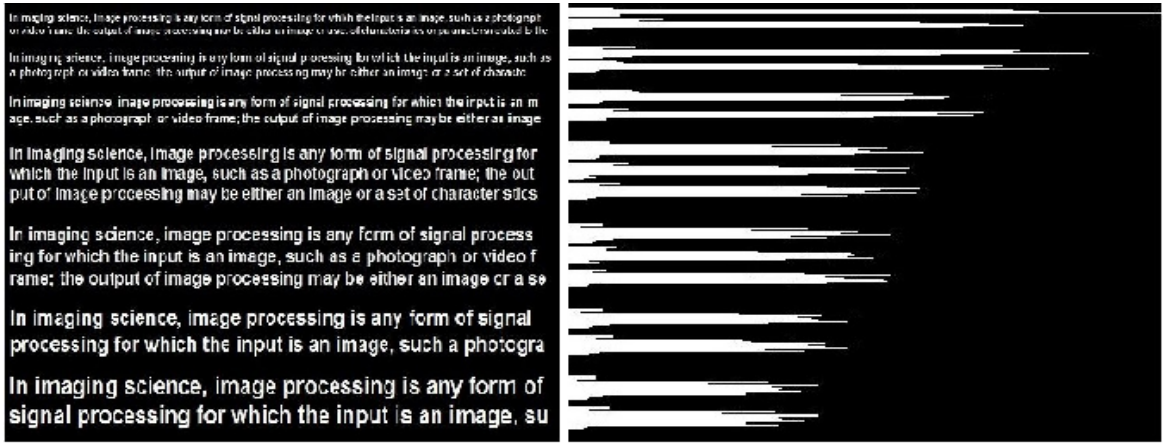

(a) Uncompressed Document           (b) Compressed Document

Fig 2: Length pattern observed from a compressed text line of different font size

*B. Density Estimation*

The number of pixels per unit area of the text line is called as pixel density. Based on the density of pixels in various regions of a compressed text line as described in Fig-1, we classify the text lines into following types: (i) ascender or descender rich (ii) both ascender and descender rich and (iii) base rich. A text line is said to be ascender or descender rich, if the descender or ascender density in the corresponding text line is close to its base density. On the other hand, the text line is base rich if both descender and ascender densities are equal to the base density. The different density parameters associated with a compressed text line are given in equations (3),(4),(5) and (6), where P(i) is the vertical projection profile[7] of the text line with length m' as given in equation (8); 'm1' and 'm2' are respective top and bottom profile of the base region; and 'r' is the length of text line in the uncompressed version.

$$\text{Line Density} = \frac{\sum_{i=1}^{m'} P(i)}{m' \times r} \quad (3)$$

$$\text{Base Density} = \frac{\sum_{i=m1}^{m2} P(i)}{b \times r} \quad (4)$$

$$\text{Ascender Density} = \frac{\sum_{i=1}^{m2} P(i)}{a \times r} \quad (5)$$

$$\text{Descender Density} = \frac{\sum_{i=m1}^{m'} P(i)}{d \times r} \quad (6)$$

The ascender and descender density parameters computed for a text line can be used to distinguish mixed case text lines from upper case text lines. Generally, the upper case text lines for a given font size have greater ascender density when compared to mixed case text lines. However, when text lines of different font size are considered, the density parameter varies non-uniformly and hence becomes difficult to generalize for font size detection. Nevertheless, for upper case text lines, the average density of profiles at P(h1) and P(h2) shown in Fig-1, also called as top and bottom profile is always greater when compared to the corresponding densities in case of mixed case text lines and experimentally it is observed that this density value is greater than 25% in upper case text lines. This average density of P(h1) and P(h2) can be used to distinguish and separate mixed case text lines from upper case text lines. This is mathematically expressed as

$$\text{MHD} = \frac{P(h1) + P(h2)}{2 \times r} \times 100 \quad (7)$$

However, in this research we aim to demonstrate the idea of font size detection with mixed case text lines. Interestingly in case of mixed case text lines, the MHD value can be used to differentiate ascender rich and both ascender and descender rich text lines. The density of P(h2) in case of ascender rich text lines is greater than density of P(h1) which results in slightly higher density when compared to ascender and descender rich text lines that have low densities at both P(h1) and P(h2). The range of MHD values obtained experimentally for different font size text lines of Arial font during training is reported in Table-1. In this research study we use MHD value to distinguish ascender rich and both ascender and descender rich text lines prior to font size recognition stage.

TABLE 1: MHD IN MIXED CASE FONT TEXT LINES

|     | Text Lines | Average Density |
|-----|------------|-----------------|
| (1) | Ascender Rich | 7% > MHD < 25% |
| (2) | Ascender & Descender Rich | MHD < 7% |
| (3) | Upper Case | MHD > 25% |

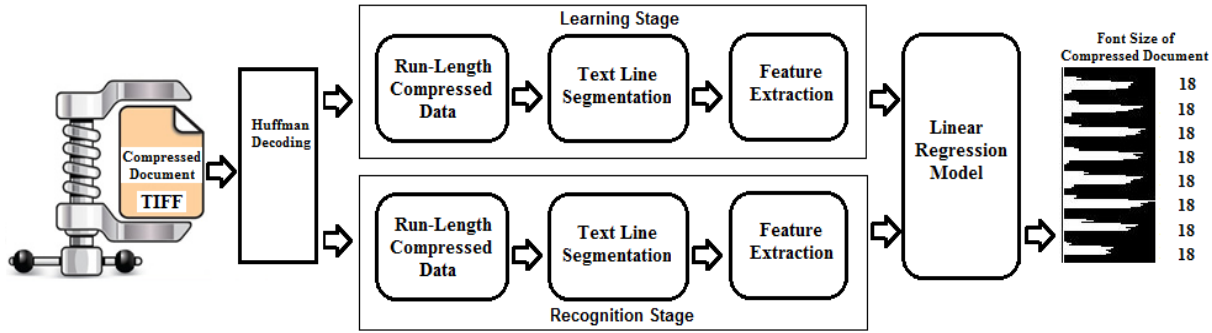

Fig 3: Proposed Model for font size detection in run-length compressed domain

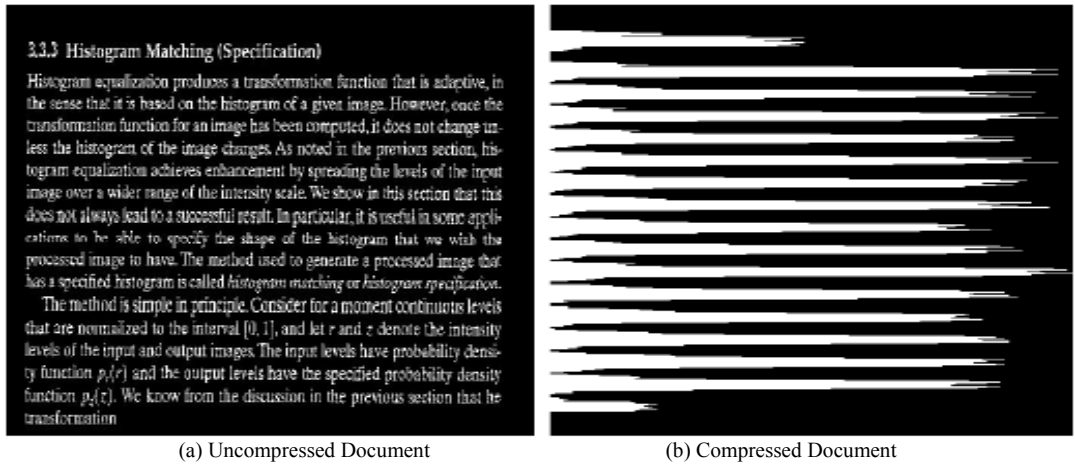

(a) Uncompressed Document  (b) Compressed Document

Fig 4: A sample document when in uncompressed and compressed versions

*C. Regression Based Learning*

In this work, we also introduce regression line based learning of font size from the text line features extracted for different font size directly from the compressed documents. During training, the knowledge of font size from the features of text lines of different font size are captured and represented in the form of a regression line y = p x + q, where p is slope of the line and q is intercept with y axis. This regression line is used to predict the font size of the text lines with the value of height features computed from the text lines of documents during testing stage. The different regression lines obtained for text line features such as line height and ascender height are discussed in section-4.

### III. PROPOSED MODEL

The proposed model shown in Fig-3 demonstrates the overall idea of automatic detection of font size in compressed documents. The model has two stages: first stage called as learning stage where the different font size are learnt from the text lines of the training set resulting in a regression line and the second stage is recognition where the unknown font size of the text lines are detected with the knowledge of regression lines. The different components involved in learning and detecting the font size from compressed documents are text segmentation, feature extraction, proper **modelling** of the features and font size detection all carried out in compressed format of a document. Here, the run-length compressed document is fed as an input to the algorithm. The underlying data structural aspects and challenges involved in handling run length compressed documents can be obtained from [7]. However in order to understand the problem better, a sample TIFF document in uncompressed and compressed versions is shown in Fig-4, where the white region in Fig-4b represents the run length compressed data of the text line.

*A. Text-Line Segmentation*

Text line segmentation in the context of a compressed document means extracting the compressed data of every text line from the compressed data of the document. The computationally efficient Vertical Projection Profile(VPP) technique proposed for compressed printed text documents in [7] is used for text line segmentation. The vertical projection profile for a compressed text line is expressed mathematically as

$$P(i) = \sum_{j=1}^{\frac{n'}{2}} b(i, j) \qquad (8)$$

The time complexity for obtaining a vertical projection profile of a compressed text line is O(m*n'/2). However in general, the time complexity of obtaining a VPP from a run length compressed document is very much lower than from its uncompressed version which has been demonstrated by [7].

*B. Feature Extraction*

After text line segmentation, using the vertical projection profile of the compressed line, a differential projection profile curve[2] which is the difference of pixel sums between adjacent scan line sums of P(i), is obtained and is used further to compute the text line features defined in Fig-1. It is mathematically expressed as

$$P'(i) = P(i+1) - P(i), \forall i = 1..m' \qquad (9)$$

In case mixed case text lines, the differential projection profile produces two peaks called as positive maxima and negative minima. Let the points at which the peaks occur be denoted as m1 and m2 respectively. They can be recognized as follows

$$m1 = i, \text{ for } \max\{P'(i)\} \ \& \ m2 = i, \text{ for } \min\{P'(i)\} \qquad (10)$$
$$\forall i = 1..m'-1$$

he distance between these two peaks give the base height(b) of the text line and the remaining features of the text line can be computed easily from this differential profile as shown in Table-2.

TABLE 2: TEXT LINE FEATURE EXTRACTION

|     | Features        | Formula      |
| --- | --------------- | ------------ |
| (1) | Line height     | h = m'       |
| (2) | Base height     | b = m2 – m1  |
| (3) | Ascender height | a = m2       |
| (4) | Descender height| d = m' - m1  |

The differential projection profile for text line 13 of Fig-2b demonstrates the extraction of these text line features as given in Fig-5. Also the vertical projection profile and differential projection profile obtained for the text lines of the entire document of Fig-2a are shown in Fig-6. More details regarding differential projection profile in the context of uncompressed text lines is available in [2]. In this research, the line height and ascender height computed from compressed text line are used for font size detection from compressed documents.

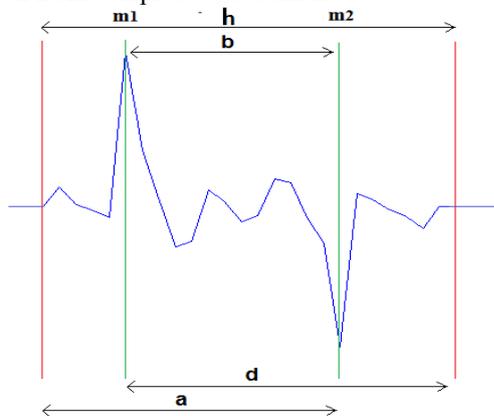

Fig 5: A differential projection profile showing extraction of text line features

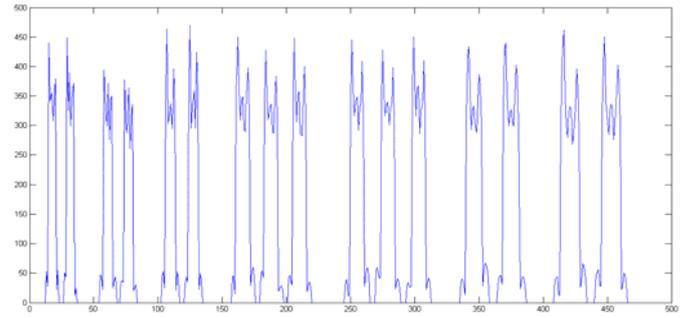
(a) VPP curve

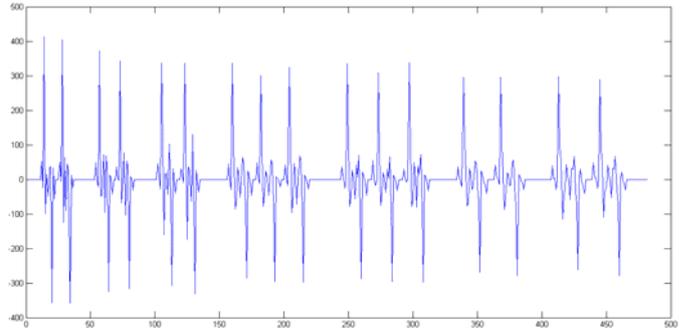
(b) Differential VPP curve

Fig 6: Vertical projection profile and its differential profile curve from a compressed document in Fig-2a

The length m' of the vertical projection profile P(i), is taken as the height of the text line. Because the font size of a text line largely depends on the height of the text line, we use height feature for font size detection. Moreover the line height feature is simple and computationally efficient which can be straight away used for font size detection. Another advantage offered by line height feature in font size detection is that it avoids computation of differential projection profile. However the line height feature in case of ascender rich text lines for a given font size is equal to ascender height of the text lines of the same font size. Therefore such text lines have to be detected using the MHD value before font size recognition and the line height computed should be resolved with regression line of ascender height feature obtained from the training document samples. Further details are discussed in section-4. The four text lines shown in Fig-7 are examples for ascender rich text lines.

which has led to drastic accretion in the volume of the data to be
is therefore useful if the documents are archived and
characteristic features of the document are known.
lines can often be seen.

Figure 7: Special case of extracted text lines demonstrating absence of descenders

The different text line features extracted using the training samples of 7 compressed documents of 7 different font sizes each containing text lines of different font size are tabulated in Table-3. In this research to detect font size of text lines, we use only line height and ascender height features. The feature values for these features in Table-3 show a variation of one pixel. Therefore while training we take the average value of these features to learn the font size of the text lines.

TABLE 3: FEATURES COMPUTED FROM THE TRAINING SET OF 7 COMPRESSED DOCUMENTS FOR 7 STANDARD FONT SIZE

| Font Size | height | base | ascender | descender |
|---|---|---|---|---|
| 8 | [32,33] | [19,20] | [26,27] | [26,27] |
| 10 | [42,42] | [22,23] | [32,33] | [32,33] |
| 12 | [48,49] | [29,30] | [38,39] | [38,39] |
| 14 | [57,58] | [32,33] | [44,45] | [44,45] |
| 16 | [60,61] | [35,36] | [48,49] | [48,49] |
| 18 | [69,71] | [41,42] | [54,55] | [57,58] |
| 20 | [79,80] | [48,48] | [63,64] | [63,64] |

During training, a regression line is used to learn the different font size of the text lines using the features extracted in Table-3. The line height and ascender height features are observed to vary linearly to the font size and therefore, we fit a linear regression and capture the pattern of font size variation. The regression lines obtained for line height and ascender height features are plotted in Fig-8. The knowledge of font size is captured in equations of regression lines in Fig-8a and Fig-8b as $y = 3.7321\ x + 3.5357$, where the norm of residuals is 3.8591 and $y = 2.9464\ x + 2.8214$, where norm of residuals is 2.719 respectively. During the testing stage, the height feature extracted for the segmented text lines are used to predict the font size of the text lines based on the knowledge of regression line. For detecting the font size of ascender and descender rich text lines, the regression line of height feature is used, while for ascender rich text lines the regression line of ascender height is used. During the detection stage, the font size that is nearest to the predicted font size value by the regression line is taken as font size of the text line.

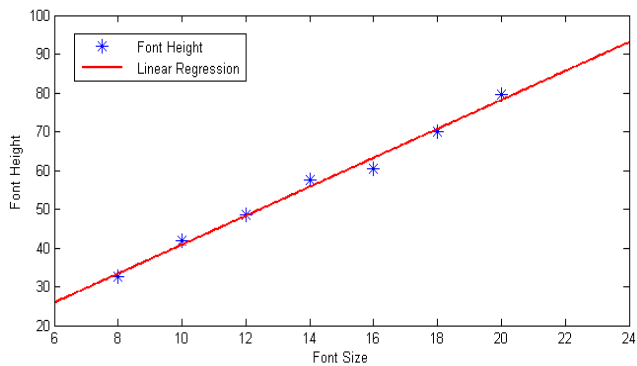
(a) Line height feature

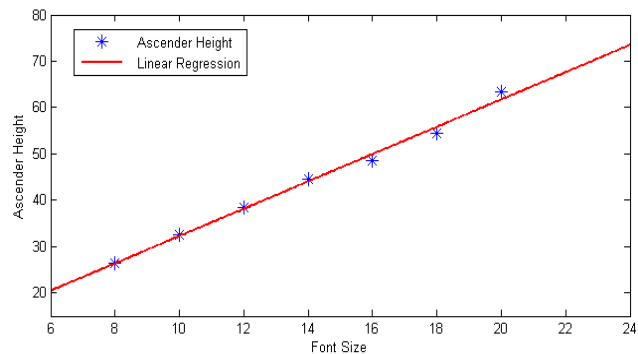
(b) Ascender height feature

Fig 8: Regression line obtained for text line height and ascender height features

## IV. EXPERIMENTAL ANALYSIS

In order to test the proposed method, two sets of data is collected consisting of totally 50 noise and skew free compressed documents (Size 2375 x 3200 at 300dpi) of 7 standard font size (8, 10, 12, 14, 16, 18 and 20) of Arial font for experimentation. The dataset-I has 35 documents which are of single font size text lines consisting of 5 documents each from every font size and the dataset-II consists of 15 documents of multiple or mixed font size text lines. The ground truth for both the data set is manually fixed and the related statistics is given in Table-4.

For training purpose 7 compressed documents are randomly chosen, one document from each font size of dataset-I. Each document is segmented into text lines, the line height and ascender height features are extracted and the knowledge of the font size is captured and preserved in the form of a regression lines. Now in order to check the performance of line height feature alone in detecting font size, we experiment with the regression line of font height with all the documents from dataset-I and dataset-II. The accuracy of font size detection obtained using the regression line of line height feature for both the data sets is tabulated in Table-5.

TABLE 4: GROUND TRUTH DETAILS OF DATASET-I & DATASET-II

| Font size | Number of text lines | |
|---|---|---|
| | Dataset-I (35 documents) | Dataset-II (15 documents) |
| 8 | 67 | 17 |
| 10 | 83 | 73 |
| 12 | 99 | 66 |
| 14 | 114 | 43 |
| 16 | 132 | 79 |
| 18 | 144 | 35 |
| 20 | 141 | 62 |
| **Total** | **780** | **375** |

TABLE 5: ACCURACY OF FONT SIZE DETECTION USING LINE HEIGHT FEATURE

| Document | Font Size | Accuracy (%) |
|---|---|---|
| Dataset-I | 8 | 98.50 |
| | 10 | 95.18 |
| | 12 | 95.96 |
| | 14 | 96.49 |
| | 16 | 93.18 |
| | 18 | 88.89 |
| | 20 | 88.65 |
| Dataset-II | 8 | 100 |
| | 10 | 98.63 |
| | 12 | 96.97 |
| | 14 | 95.35 |
| | 16 | 93.67 |
| | 18 | 82.86 |
| | 20 | 93.55 |

Fig 9: Text lines demonstrating the presence of numbers, special symbols, italic words and numbers

With the above experiment it is proved that line height can be potentially a good feature for font size detection of text lines in compressed documents. Moreover at line level, this feature can also handle italics fonts, numbers and special symbols embedded in the text line, which is shown Fig-9. However, the performance obtained with the feature is not satisfactory but could be improved using an additional ascender height feature in the recognition stage. In case of line height feature, the decrease in performance for most of the time is observed due to ascender rich text lines which do not include any descender character in them. In such text lines the ascender height is wrongly taken as line height, and hence the predicted font size is incorrect. In order to improve the performance of font size detection with line height feature, the ascender rich text lines need to be distinguished and recognized separately with its ascender height feature. Therefore, we use MHD value to classify the ascender rich text lines from ascender and descender rich text lines. For ascender rich text lines, the font size is predicted with the knowledge of regression line obtained from ascender height features during the training stage. The accuracy of font size recognition using line height and ascender height as an additional feature on dataset-I & dataset-II is given in Table-6.

TABLE 6: ACCURACY OF FONT SIZE DETECTION USING LINE HEIGHT AND ASCENDER HEIGHT FEATURES

| Document | Font Size | Accuracy (%) |
|---|---|---|
| Dataset-I | 8 | 100 |
| | 10 | 98.79 |
| | 12 | 100 |
| | 14 | 100 |
| | 16 | 99.24 |
| | 18 | 100 |
| | 20 | 100 |
| Dataset-II | 8 | 100 |
| | 10 | 98.63 |
| | 12 | 100 |
| | 14 | 100 |
| | 16 | 98.70 |
| | 18 | 100 |
| | 20 | 100 |
| Overall | | 99.67 |

Text line font size detection with ascender height as an additional feature produces commendable results. However, during the experiment we observe some small error which may appear in the last line of a paragraph. Such error occurs in a text line containing few words which may belong to a category of base rich text line. The height feature computed in these text lines are less than the actual height of ascender rich text lines and hence leads to recognition error. In case such text line occurs, it is possible to detect their font size with MHD value and base height regression line. Nevertheless, occurrence of such error in text lines of sufficient length is an unlikely event[2]. Therefore, this issue has not been addressed in this research work. Two test documents, one each from dataset-I & dataset-II showing the detection of font size from compressed documents are given in Fig-10.

Finally, we also extrapolate the knowledge of font size in the form regression line to predict the font size of text lines (9, 11, 13, 15, 17 and 19) which were not the part of training set. In the similar manner, the line height and ascender height are computed from the segmented text lines and their font size is predicted with the regression line. The result of font size detection obtained for the test documents considered which are decompressed and shown in Fig-11. We observe that font size 12 & 13, 15 & 16 and 17 & 18 have line height in the same range and hence results have been wrongly predicted as highlighted and shown in Fig-11. Other than these fonts, the prediction of font size with the proposed model works satisfactorily.

(a) Test document-22

(b) Test document-50

Figure 10: Results for automatic font size in documents of similar font size and mixed font size text lines

The overall goal of this research work was to demonstrate the idea of font size detection directly in compressed documents. Therefore, the focus of font size detection was limited to line level and single font style (Arial). However, the same idea of font size detection could also be extended to word level and with different font styles which could be taken as an extension work to this research study.

## V. CONCLUSION

In this research work, a novel method for automatic detection of font size directly from the run length compressed documents at line level is proposed. The simple text line features such as line height and ascender height are computed from the compressed text line of different font size and this knowledge is preserved in the form a regression line which is used to predict the font size during the testing stage. Further, the regression line is also experimented to predict the font size of text lines which were not considered for training. Overall, the idea presented for font size detection in run length compressed domain is validated with sufficient data set and the performance results obtained are commendable enough to motivate to extend the current idea to word level and also with other font styles.

(a) Test document-51

(b) Test document-52

Figure 11: Results for automatic font size prediction for untrained text line fonts